\documentclass{article}

\usepackage[final]{nips_2017}
\usepackage{xcolor}
\usepackage{tikz}
\usepackage{amssymb}
\usetikzlibrary{tikzmark, positioning, arrows}

\usepackage[utf8]{inputenc} % allow utf-8 input
\usepackage[T1]{fontenc}    % use 8-bit T1 fonts
\usepackage{hyperref}       % hyperlinks
\usepackage{url}            % simple URL typesetting
\usepackage{booktabs}       % professional-quality tables
\usepackage{amsfonts}       % blackboard math symbols
\usepackage{nicefrac}       % compact symbols for 1/2, etc.
\usepackage{microtype}      % microtypography
\usepackage{wrapfig}

\usepackage{bbm}  % indicator function
\usepackage{amsmath}
\usepackage{graphicx}
\usepackage{makecell}
\usepackage{multirow}
\usepackage{changepage}

\usepackage{enumitem}
\usepackage{flafter}

\title{Proportionate gradient updates with PercentDelta}

\author{
   Sami Abu-El-Haija \\
  Google Research \\
  Mountain View, CA \\
  \texttt{haija@google.com} \\
}

\begin{document}

\maketitle

\begin{abstract}
Deep Neural Networks are generally trained using iterative gradient updates. Magnitudes of gradients are affected by many factors, including choice of activation functions and initialization. More importantly, gradient magnitudes can greatly differ across layers, with some layers receiving much smaller gradients than others. causing some layers to train slower than others and therefore slowing down the overall convergence. We analytically explain this disproportionality. Then we propose to explicitly train all layers at the same speed, by scaling the gradient w.r.t. every trainable tensor to be proportional to its current value. In particular, at every batch, we want to update all trainable tensors, such that the relative change of the L1-norm of the tensors is the same, across all layers of the network, throughout training time. Experiments on MNIST show that our method appropriately scales gradients, such that the relative change in trainable tensors is approximately equal across layers. In addition, measuring the test accuracy with training time, shows that our method trains faster than other methods, giving higher test accuracy given same budget of training steps.
\end{abstract}

\section{Introduction}
Weights of Deep Neural Networks (DNNs) are commonly randomly initialized. The $j$-th layer's weight matrix $W_j$ is initialized from a distribution that is generally conditioned on the shape of $W_j$.

Feed-forward DNNs generally consist of a series of matrix multiplications and element-wise activation functions. For example, input vector $\mathbf{x_0}$ is transformed to the output vector $\mathbf{x}_L$ by an $L$-layer neural network $h_\theta(\mathbf{x_0}) =  \mathbf{x}_L$, depicted as:

\begin{tikzpicture}
\node (x0) []{$\mathbf{x}_0$};
\node (w0) [above right=of x0]{$W_0$};
\node (m0) at (x0 -| w0) [draw,thick,minimum width=0.5cm,minimum height=0.5cm] {$\times$}
edge [<-]  node [] {} (w0)
edge [<-]  node [] {} (x0);
\node (a0) [draw,thick,minimum width=0.5cm,minimum height=0.5cm, right=of m0] {$\sigma$}
edge [<-]  node [] {} (m0);

\node (w1) [above right=of a0]{$W_1$};
\node (m1) at (a0 -| w1) [draw,thick,minimum width=0.5cm,minimum height=0.5cm] {$\times$}
edge [<-]  node [] {} (w1)
edge [<-]  node [] {} (a0);
\node (a1) [draw,thick,minimum width=0.5cm,minimum height=0.5cm, right=of m1] {$\sigma$}
edge [<-]  node [] {} (m1);

\node (dots) [right=of a1] {\dots}
	edge [<-] node [] {} (a1);
\node (wl) [above right=of dots]{$W_{L-1}$};
\node (ml) at (dots -| wl) [draw,thick,minimum width=0.5cm,minimum height=0.5cm] {$\times$}
edge [<-]  node [] {} (wl)
edge [<-]  node [] {} (dots);
\node (al) [draw,thick,minimum width=0.5cm,minimum height=0.5cm, right=of ml] {$\sigma$}
	edge [<-]  node [] {} (ml);

\node (xl) [right=of al]{$\mathbf{x}_L$}
	edge [<-]  node [] {} (al);
\end{tikzpicture}

Where $\times$ is a matrix-multply operator, $\sigma$ is an element-wise activation (e.g. logistic or ReLu). The DNN parameters $\theta = \{W_0, W_1, \dots, W_{L-1}\}$ are optimized by a training algorithm, such as Stochastic Gradient Descent (SGD), which iteratively applies gradient updates:
\begin{equation}
W_j^{(t+1)} := W_j^{(t)} - \eta \ \gamma(t) \ \frac{\partial J}{\partial W_j},
\end{equation}
where $W_j^{(t)}$ is the value of the $j$-th layer weight matrix $W_j$ at timestep $t$, $\eta \in \mathbb{R}$ is the learning rate, the decay function $\gamma : \mathbb{R} \rightarrow [0, 1]$, generally decreases with time $t$, and $\frac{\partial J}{\partial W_j}$ is the partial gradient of training objective $J$ w.r.t. $W_j$, evaluated at time $t$ for a batch of training examples. For notational convenience, we define the delta SGD as:  $\Delta_{\text{SGD}} W =  \eta \ \gamma(t) \ \frac{\partial J}{\partial W}$

%The gradients of the training objective $J$ w.r.t. output vector $\mathbf{x}_L$ is defined as $\frac{\partial J}{\partial \mathbf{x}_L}$. The gradients w.r.t. weight matrices $\{W_0, W_1, \dots, W_{L-1}\}$ is generally calculated using the Back Propagation Algorithm \citep[BackProp, ][]{backprop}.

Under usual circumstances, the magnitudes of gradients can widely vary across layers of a DNN, causing some trainable tensors to change much slower than others, slowing down overall training. Several proposed methods metigate this problem, including utilizing: per-parameter adaptive learning rate such as AdaGrad \citep{adagrad} or Adam \citep{adam}; normalization operators such as BatchNorm \citep{batchnorm}, WeightNorm \citep{weightnorm}, or LayerNorm \citep{layernorm}; and intelligent initialization schemes such as xavier's \citep{xavier}. These methods heuristically attack the disproportionate training problem, which we justify in section \ref{sec:motivation}. In this paper, we propose to directly enforce proportionate training of layers. Specifically, We propose a gradient update rule that moves every weight tensor in the direction of the gradient, but with a magnitude that changes the tensor value by a relative amount. We use the same relative amount across all layers, to train them all at the same speed.

The remainder of the paper is organized as follows. In Section \ref{sec:motivation} we illustrate the disproportionate training phenomena on a (toy) hypothetical 4-layer neural network, by expanding gradient terms using BackProp \cite{backprop}.  In Section \ref{sec:related}, we summarize related work. Then, we introduce our algorithm in Section \ref{sec:pd}. We show experimental results on MNIST in Section \ref{sec:experiments}.  In Section \ref{sec:discussion}, we discuss where PercentDelta could be useful and potential future direction. Finally, we conclude our findings in Section \ref{sec:conclusion}.

\section{Disproprtionate Training}
\label{sec:motivation}
We use a toy example to illustrate disproportionate training across layers. Assume a 4-layer network with trainable weight matrices $\{W_0, W_1, W_2, W_3\}$ and output vector $\mathbf{x}_4$. The gradient of the objective w.r.t. output vector $\mathbf{x}_4$ can be directly calculated from the data e.g. using cross-entropy loss. We write-down the gradient of the objective $J$ w.r.t. the last layer's weight matrix $W_3$ as:
\begin{equation}
\frac{\partial J}{\partial W_3} = \left[\sigma'(W_3 \times \mathbf{x}_3)  \circ \frac{\partial J}{\partial \mathbf{x}_4} \right] \times \mathbf{x}_3^T,
\end{equation}
where $\sigma'()$ is the derivative of $\sigma()$ w.r.t. its input, and $\circ$ is the Hadamard product.

We also write-down the gradient w.r.t. $W_1$ and $W_2$, then expand the expressions using the Back Propagation Algorithm \citep{backprop}:
\begin{align*}
\frac{\partial J}{\partial W_2} &= \left[\sigma'(W_2 \times \mathbf{x}_2)  \circ \underbrace{\frac{\partial J}{\partial \mathbf{x}_3}}_{=\tikzmark{s1}} \right] \times \mathbf{x}_2^T \\
  &= \left[\sigma'(W_2 \times \mathbf{x}_2)  \circ \overbrace{\left[ W_3^T \times \left[ \sigma'(W_3 \times \mathbf{x}_3)  \circ \frac{\partial J}{\partial \mathbf{x}_4}  \right] \right]}^{\tikzmark{t1}} \right] \times \mathbf{x}_2^T \\
\frac{\partial J}{\partial W_1} &= \left[\sigma'(W_1 \times \mathbf{x}_1)  \circ \underbrace{\frac{\partial J}{\partial \mathbf{x}_2}}_{=\tikzmark{s2}} \right] \times \mathbf{x}_1^T \\
&= \left[\sigma'(W_1 \times \mathbf{x}_1)  \circ \overbrace{\left[ W_2^T \times \left[ \sigma'(W_2 \times \mathbf{x}_2)  \circ \underbrace{ \left[ W_3^T \times \left[ \sigma'(W_3 \times \mathbf{x}_3)  \circ \frac{\partial J}{\partial \mathbf{x}_4}  \right] \right]}_{=\partial J / \partial \mathbf{x}_3}  \right] \right]}^{\tikzmark{t2}} \right] \times \mathbf{x}_1^T
\end{align*}
\begin{tikzpicture}
\tikz[remember picture]
\draw[-{>[scale=1.0]}, overlay] (pic cs:s1)to  (pic cs:t1);
\end{tikzpicture}
\begin{tikzpicture}
\tikz[remember picture]
\draw[-{>[scale=1.0]}, overlay] (pic cs:s2)to  (pic cs:t2);
\end{tikzpicture}

Note the following in the above equations: First, the derivatives $\frac{\partial J}{\partial W_3}, \frac{\partial J}{\partial W_2}, \frac{\partial J}{\partial W_1}$ look similar. In fact, they are \textit{almost} sub-expressions of one another, with the exception of the right-most row-vector $\mathbf{x}_j^T$. Second, all quantities in brackets are column-vectors. The gradient matrix $\frac{\partial J}{\partial W_j}$ is determined by an outer product of a column-vector (in brackets) times a row-vector $\mathbf{x}_j^T$.

What can we conclude about the magnitudes (e.g. the L1 norm) of $\frac{\partial J}{\partial W_3}, \frac{\partial J}{\partial W_2}, \frac{\partial J}{\partial W_1}$? Each  $\frac{\partial J}{\partial W_j}$ is calculated using three types of multiplicands: $\sigma'(.)$, $W_.^T$, and $\mathbf{x}_j^T$. Therefore, the magnitudes of $\frac{\partial J}{\partial W_j}$ are affected by:
\begin{enumerate}
	\item The value of the derivative of $\sigma'()$, which is evaluated element-wise. If $\sigma'()$ is generally less than 1, then we can expect the gradient to be smaller for earlier layers than later ones, since they are multiplied by $\sigma'()$ more times. If it is generally greater than 1, then we can expect the gradient to be larger for earlier layers. For very deep networks (recurrent or otherwise), the former situation can cause the gradients to \textit{vanish}, while the latter can cause the gradients to \textit{explode}. ReLu mitigates this problem, as its derivative $= 1$, in locations where input is positive.

	\item The magnitude of $W$'s. In practice, all $W$'s are initialized from the same distribution. If the L1-norms of rows in $W_j^T$ are less than 1, then we should expect $||W_j^T \times \sigma'(\mathbf{z})|| < ||\sigma'(\mathbf{z})||$, yielding smaller gradients for earlier layers. Otherwise, if the row L1 norms are greater than 1, then earleir layers should receive larger gradient magnitudes. Weight normalization \citep{weightnorm} and intelligent initialization schemes \citep[e.g.][]{xavier} can mitigate this problem.
	
	\item The norm of the row-vector $\mathbf{x}_j$. If in the forward pass, the activations consistently grow (e.g. unbounded activation) with subsequent layers, then later layers will receive larger gradients. BatchNorm \citep{batchnorm} and LayerNorm \citep{layernorm} mitigate this problem.
\end{enumerate}

%, are learned using a training algorithm. The parameter values, and the rate at which they change during training are determined by factors including choice of activation function and initialization.
%function dependent on many factors , the values of the parameters change to meet an objective. The rate at which they change and their values is affected by  change . Trate Initialization of parameters, choice of activation function,

\section{Related Work}
\label{sec:related}
\subsection{Adaptive Gradient}
\cite{adagrad} proposed AdaGrad. A training algorithm that keeps a cumulative sum-of-squared gradients (i.e. second moment of gradients):
\begin{equation}
S^{(t)}_j = S^{(t - 1)}_j + \left(\frac{\partial J}{\partial W_j} \circ \frac{\partial J}{\partial W_j} \right),
\end{equation}
Then divides (element-wise) the gradient by the square-root of the sum:
\begin{equation}
W_j^{(t+1)} := W_j^{(t)} - \eta \ \gamma(t) \ \left( \frac{\partial J}{\partial W_j} \circ \left(S^{(t)}_j\right)^{-1/2} \right),
\end{equation}
or equivalently $\Delta_{\textrm{AdaGrad}}W_j = \eta \ \gamma(t) \ \frac{\partial J}{\partial W_j} \circ \left(S^{(t)}_j\right)^{-1/2}$, where the $(.)^{-1/2}$ power operator is applied element-wise. In essense, if some layer receives large gradients, then they will be normalized to smaller gradients through the division. However, a weakness in AdaGrad is that at some point, the $S$ will grow too large, effectively making $(S)^{-1/2} \approx 0$ and therefore slowing or halting training.

\cite{adam} propose Adam, which keeps exponential decaying sums, of gradients and of square gradients, respectively known as first and second moments. Adam offers two benefits over AdaGrad. First, its decaying sums should not grow to infinity and therefore training should not halt. Second, the exponential-decay averaging was shown to speed up training \citep[Momentum,][]{momentum}. For details, we point readers to the Adam paper \citep{adam}.

\subsection{LARS}
\cite{lars} propose to normalize every layer's gradients by the ratio of L2 norms of the parameter and the gradient. Namely, they propose the gradient update rule:
\begin{equation}
\Delta_{\textrm{LARS}}  W_j = \eta \ \gamma(t) \ \left( \frac{\left|\left|W_j^{(t)}\right|\right|_2}{\ \left|\left|(\partial J / \partial W_j)\right|\right|_2}\right) \ \frac{\partial J}{\partial W_j},
\end{equation}
which effectively normalizes the gradient $\frac{\partial J}{\partial W_j}$ to be unit-norm. This setup is very similar to ours, with two differences: First, our norm operator is L1 rather than L2. Second, our norm operator is applied outside the division (i.e. our division is element-wise). Our proposed algorithm, PercentDelta, was used to train our work \citep{asymproj}, before we where aware of the work of \citep{lars}. In \citep{asymproj}, PercentDelta gave us $1\%$ improvement over Adam, over all datasets. Nonetheless, we only discuss MNIST experiments in this paper, and leave graph embedding experiments for follow-up work.

%\subsection{Meta-Learning}
%The goal of Meta-learning (a.k.a. Learning to Learn) is to make a fully automated training algorithm, hopefully requiring no manual hyperparameter searches. Among the wealth of recent Meta-learning works, we draw attention to \citep{joshicml}, as one of their features is \textbf{Complete}.

\section{PercentDelta}
\label{sec:pd}
For every weight matrix $W_j$, and similarily bias vectors, we propose the gradient update rule:
\begin{equation}
\label{eq:pd}
\Delta_{\textrm{PercentDelta}}  W_j = \eta \ \gamma(t) \ \left(\frac{\textrm{size}(W_j)}{\ \left|\left|\frac{(\partial J / \partial W_j)}{W_j^{(t)}}\right|\right|_1}\right) \ \frac{\partial J}{\partial W_j},
\end{equation}
where scalar $\textrm{size}(W) \in \mathbb{Z}^+$ is the number of entries\footnote{$\textrm{size}(W)$ is the product of $W$'s dimensions. For a 2-D matrix, it is equal to  \# rows $\times$ \# columns, for a vector, it is equal to its length, etc. We also define $\textrm{size(scalar)} = 1$.} in $W$, and the scalar:
\begin{equation}
\label{eq:multiplier}
\frac{\textrm{size}(W_j)}{\ \left|\left|\frac{(\partial J / \partial W_j)}{W_j^{(t)}}\right|\right|_1}
\end{equation}
normalizes the gradient $\frac{\partial J}{\partial W_j}$ of the $j$-th layer, so that it is more proportional to its current parameter value $W_j^{(t)}$, and $||.||_1$ is the L1-norm. We avoid division-by-zero errors by adding an epsilon to the denominator\footnote{In practice, rather than converting $\frac{a}{b}$ to $\frac{a}{b + \epsilon}$, we use $\frac{a}{b + \epsilon \cdot sign(b)}$, as our goal is to push $b$ slightly away from zero while keeping its sign.}. The divide operator within the L1 norm is applied element-wise and the outer divide operator is scalar.
Note that the ``gradient multiplier'' fraction (Equation \ref{eq:multiplier}) is $\in \mathbb{R}^+$. Therefore, we only changes the gardient's magnitude, but not its direction.
%multipliers[v] = target_pd / PlusEpsilon(
%    tf.reduce_mean(tf.abs(g / PlusEpsilon(v))))

Now, we justify Equation \ref{eq:pd}. Assume that $W_s$ is scalar. Therefore, $\textrm{size}(W_s) = 1$ and the L1-norm becomes an absolute value. Equation \ref{eq:pd} simplifies to:
\begin{align}
\begin{split}
W_s^{(t+1)} &:= W_s^{(t)} - \eta \ \gamma(t) \ \ \left|\frac{W_s^{(t)}}{(\partial J / \partial W_s)}\right| \ \frac{\partial J}{\partial W_s} \\
  &:= W_s^{(t)} - \eta \ \gamma(t) \ \ \left|W_s^{(t)}\right| \ \textrm{sign}\left(\frac{\partial J}{\partial W_s}\right),
\end{split}
\end{align}
where $\textrm{sign}(z) = 1$ if $z>0$ and $=-1$ if $z < 0$. Therefore, $W_s$ will change with a quantity \textit{proportional} to its current value. In particular, the scalar $\eta \ \gamma(t)$ determines the percentage at which $W$ changes at timestep $t$, giving rise to the name: PercentDelta. For example, if we setup a decay schedule on $\gamma(t)$, such that $\eta \gamma(t)$ changes during training from $10\%$ to $0.01\%$, then the network parameters will change at a rate of $10\%$ in early mini-batches, and gradually decrease their PercentDelta updates to $0.01\%$ towards end of training, consistently across all layers, regardless of the current parameter value or the gradient value, which are influenced by choice of activation function, initialization, and network architecture.

\section{Experiments}
\label{sec:experiments}
We run experiments on MNIST. We use the same model for all experiments and fix the batch size to 500. Our model code is copied from the TensorFlow tutorial\footnote{\url{https://www.tensorflow.org/get_started/mnist/pros}}, and contains 4 trainable layers: 2D Convolution, Max-pooling, 2D Convolution, Max-pooling, Fully-connected, Fully-connected. The convolutional layers contain trainable tensors with dimensions: (5, 5, 1, 32), (5, 5, 32, 64), and bias vectors. The Fully-connected layers contain trainable tensors with dimensions: (3136, 1024), (1024, 10), and bias vectors. The model is trained with Softmax loss, uses ReLu for hidden activations, and does not use BatchNorm.

\subsection{MNIST Network Gradient Magnitudes}
We record the gradient magnitudes for our 4-layer MNIST network throughout training. We compare the relative magnitude that $W$ changes, under vanilla SGD versus under PercentDelta. We plot $\left(\left|\left|\Delta_\textrm{SGD} W\right|\right|_1 / \left|\left| W \right|\right|_1 \right)$ and $\left(\left|\left|\Delta_\textrm{PercentDelta} W\right|\right|_1 / \left|\left| W \right|\right|_1 \right)$ for tranable tensors $W$ of convolutional and fully-connected layers. We notice that vanilla SGD proposes gradients that are not proportional to current weights. For example, using some learning rate, some layers can completely diverge (e.g. all entries switching signs) while other layers would change at a rate of <1\%. In this case, to prevent divergence in any layer tensors, the learning rate would be lowered. However, using PercentDelta, relative magnitude of parameter updates is almost equal for all layers, showing that all layers are training at the same speed, consistently throughout the duration of training.

\begin{figure*}[ht]
\includegraphics[width=\columnwidth]{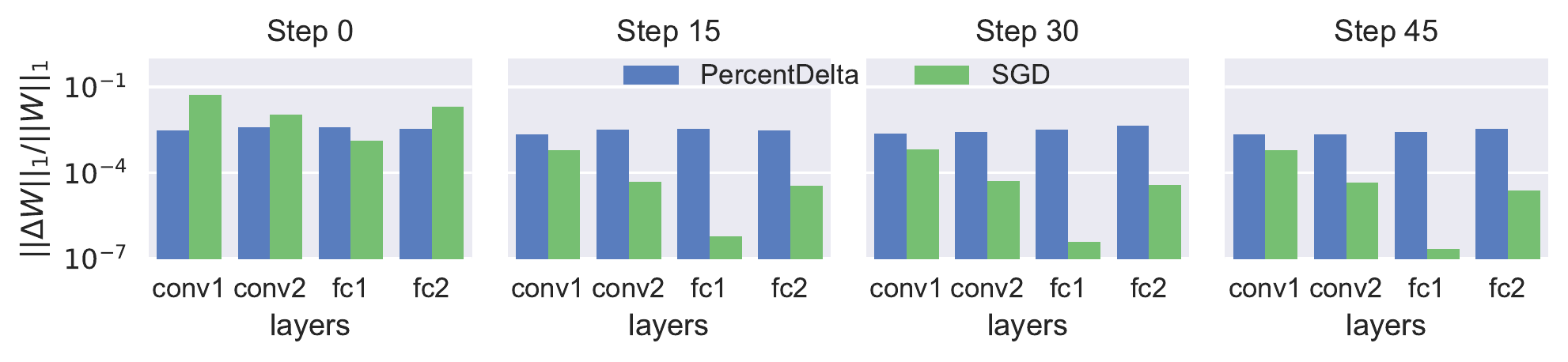}
\label{fig:mnistgrads}
\caption{Magnitude of relative deltas for every layer, at training timestamps (batch at step 0, step 15, \dots). Y-axis is the relative change of the L1 norm of every layer's trainable tensor. For Stochastic Gradient Descent (SGD) and PercentDelta, respectively, we measure it as $\left(\left|\left|\Delta_\textrm{SGD} W\right|\right|_1 / \left|\left| W \right|\right|_1 \right)$ and $\left(\left|\left|\Delta_\textrm{PercentDelta} W\right|\right|_1 / \left|\left| W \right|\right|_1 \right)$. Disproportionate training is played down by the log-scale. Without log-scale, some SGD bars wont be visible. Curves are scaled by $\eta \ \gamma(t)$.}
\end{figure*}

\subsection{MNIST Test Accuracy Curves}
We want to measure how fast can PercentDelta train MNIST. We compare training speed with other algorithms, including per-parameter adaptive learning rate algorithms, AdaGrad \citep{adagrad} and Adam \citep{adam}, as well as a recent algorithm with similar spirit, LARS \citep{lars}, which also normalizes the gradient w.r.t. a weight tensor, by the current value of the weight tensor.

\begin{figure*}[]
\begin{tabular}{c c c}
	\includegraphics[width=0.33\columnwidth]{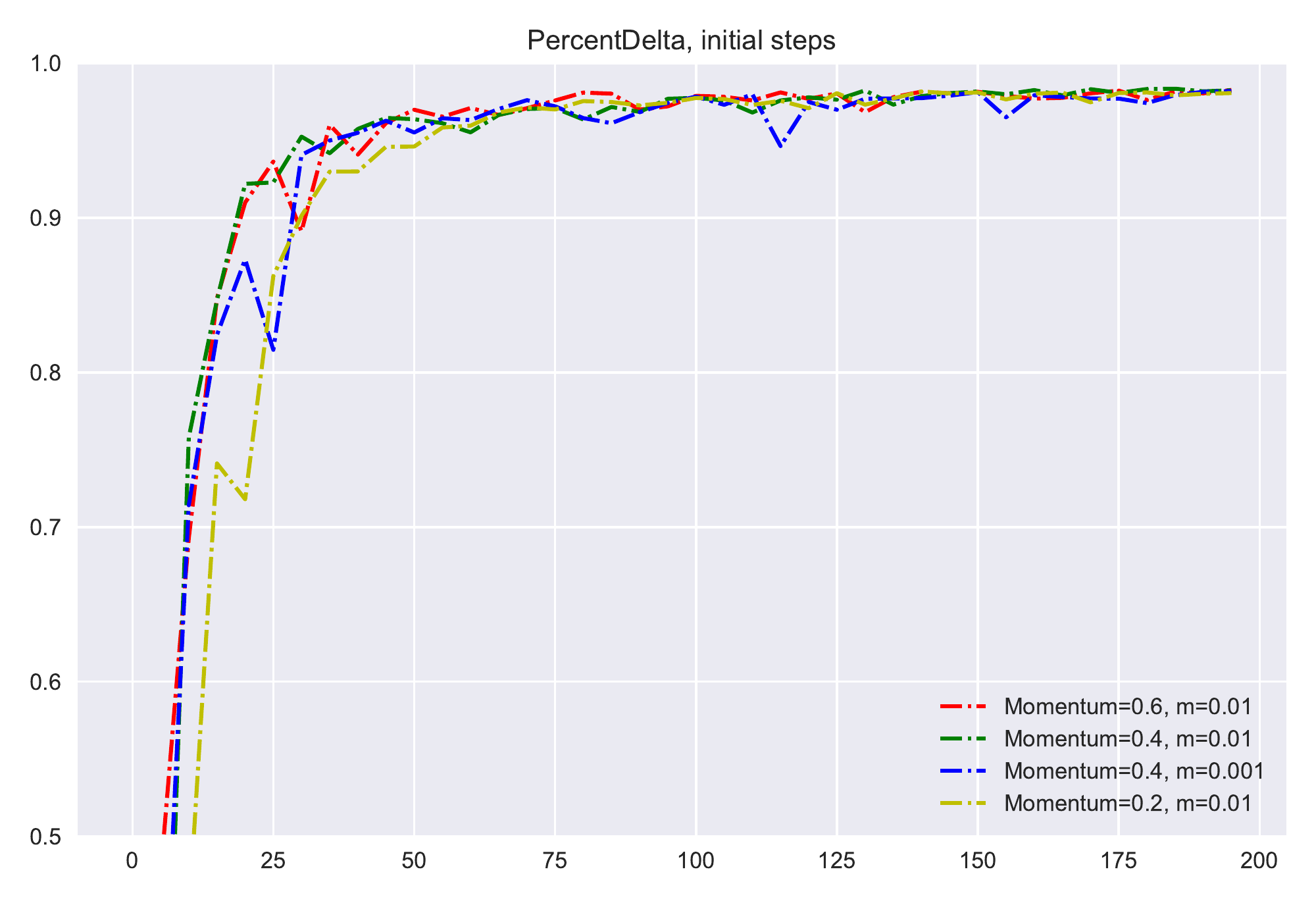}  & 	\includegraphics[width=0.33\columnwidth]{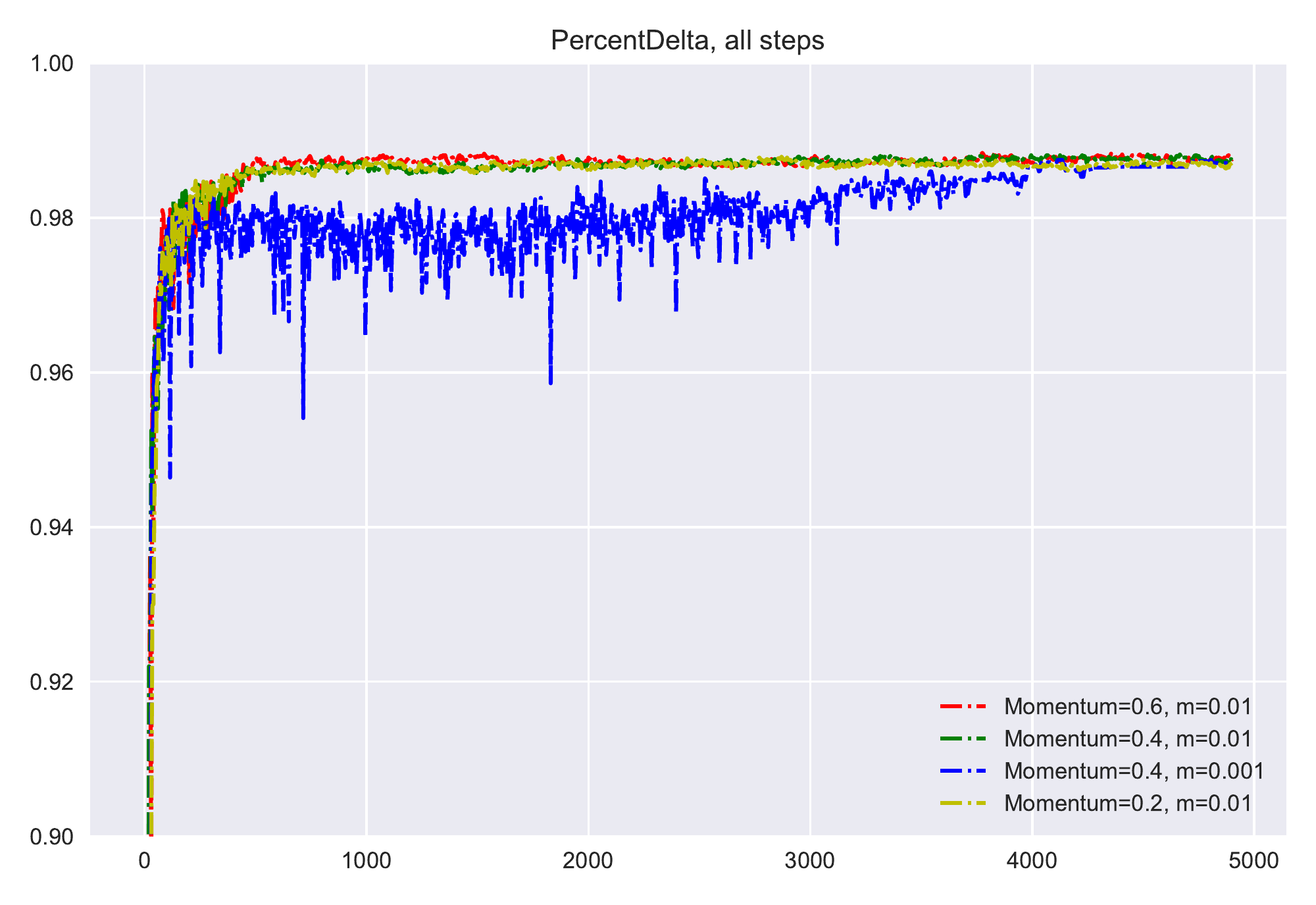}  &
	\includegraphics[width=0.33\columnwidth]{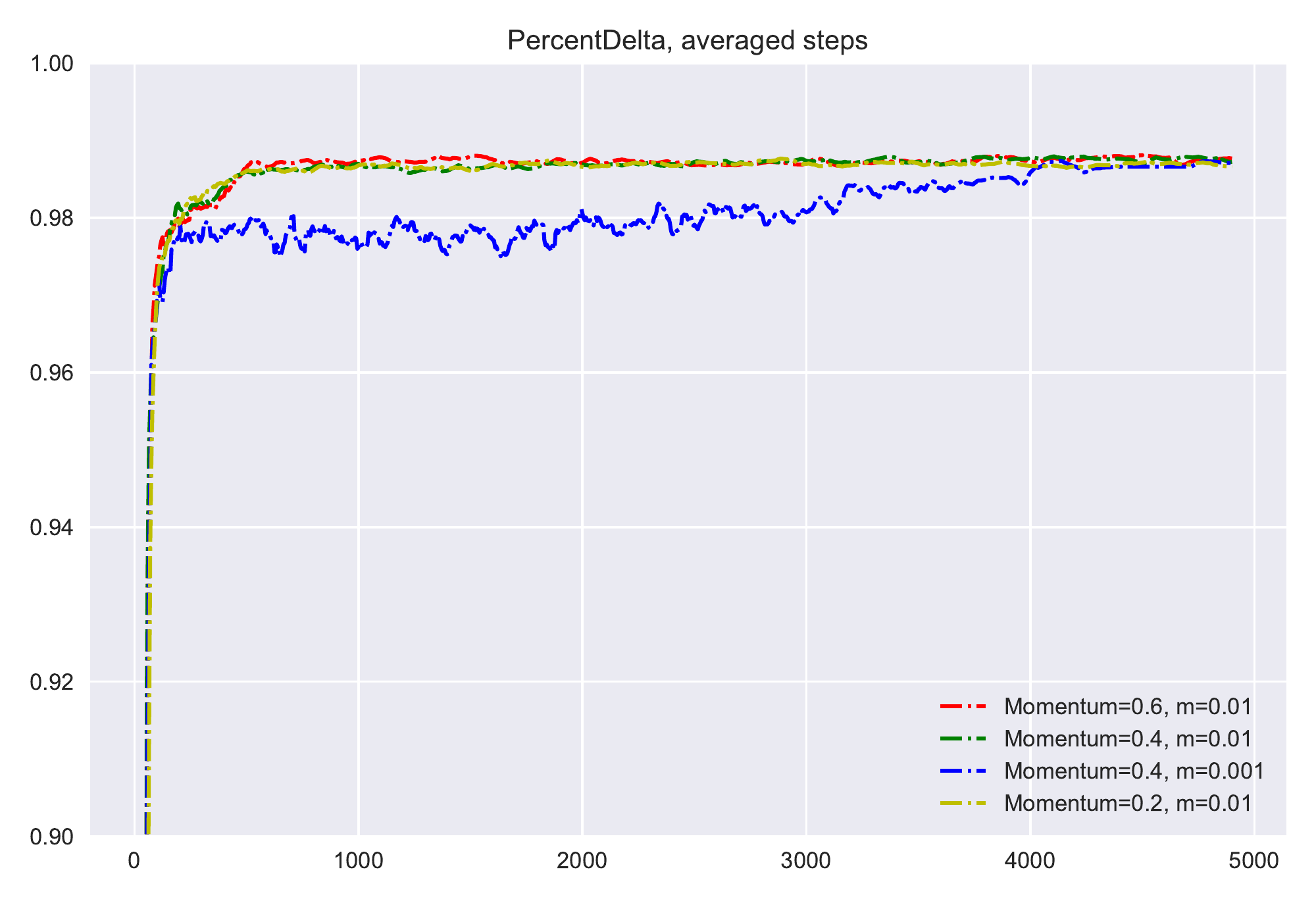}
	\\
	\includegraphics[width=0.33\columnwidth]{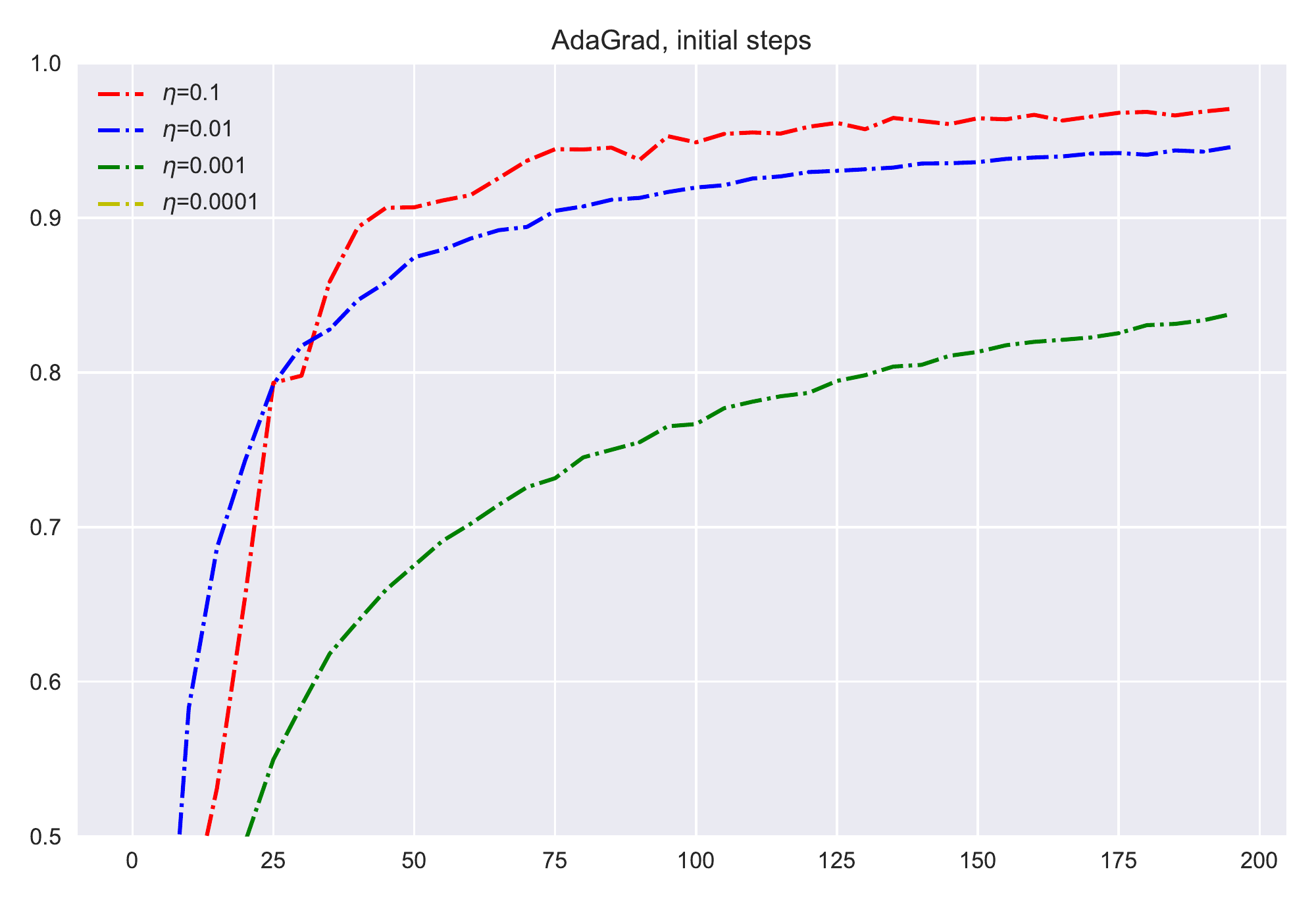}  & 	\includegraphics[width=0.33\columnwidth]{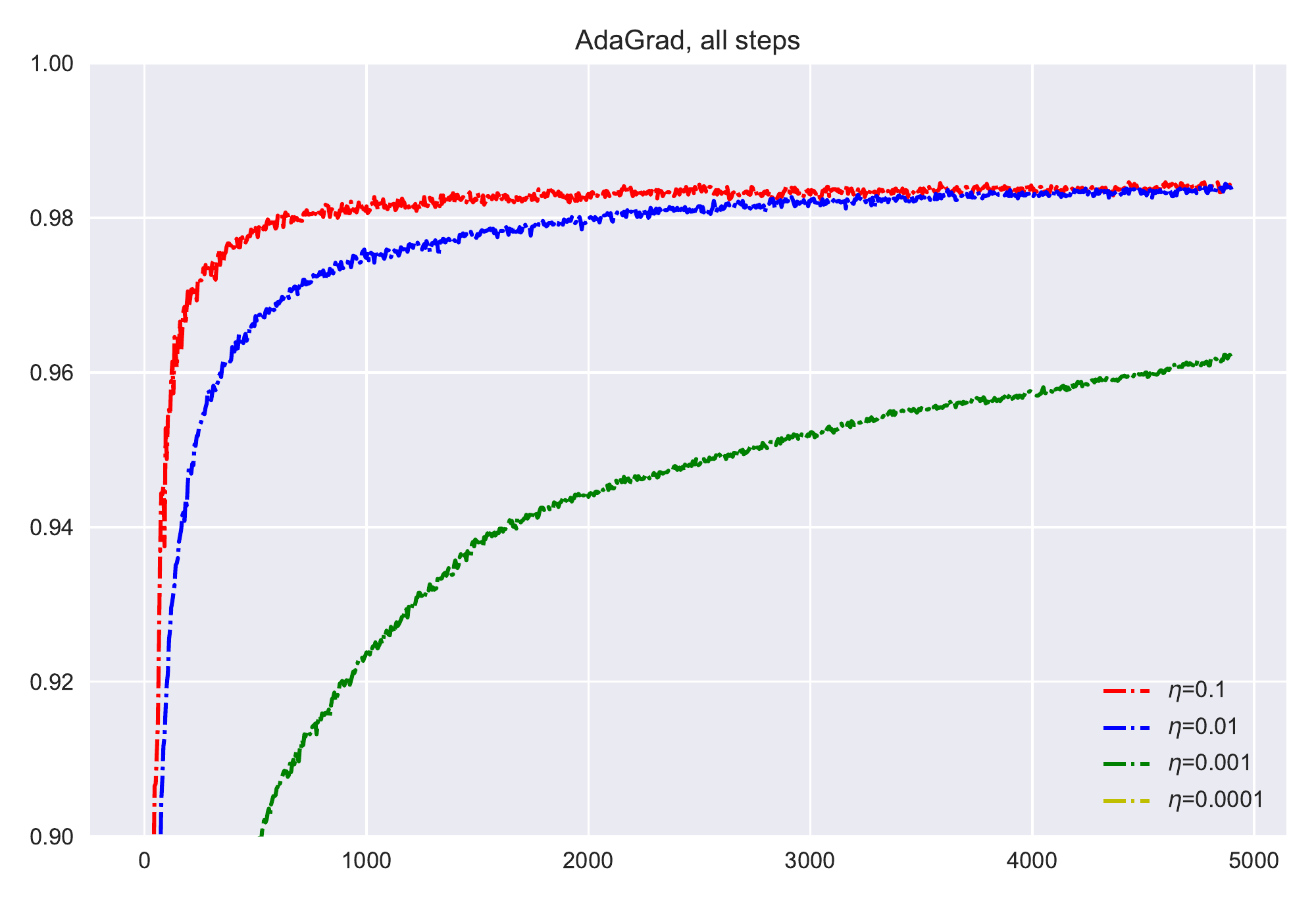}  &
	\includegraphics[width=0.33\columnwidth]{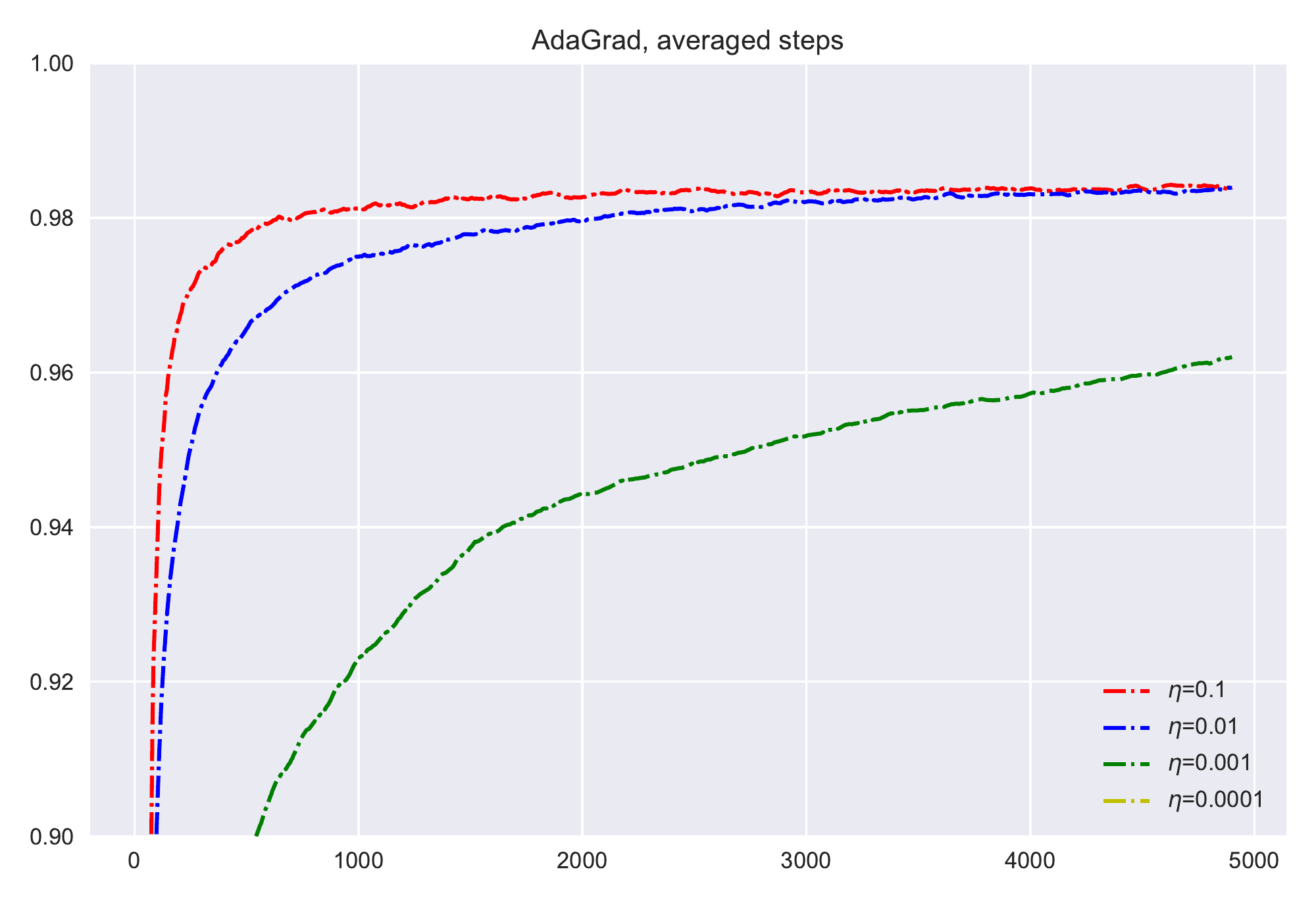}
	\\
	\includegraphics[width=0.33\columnwidth]{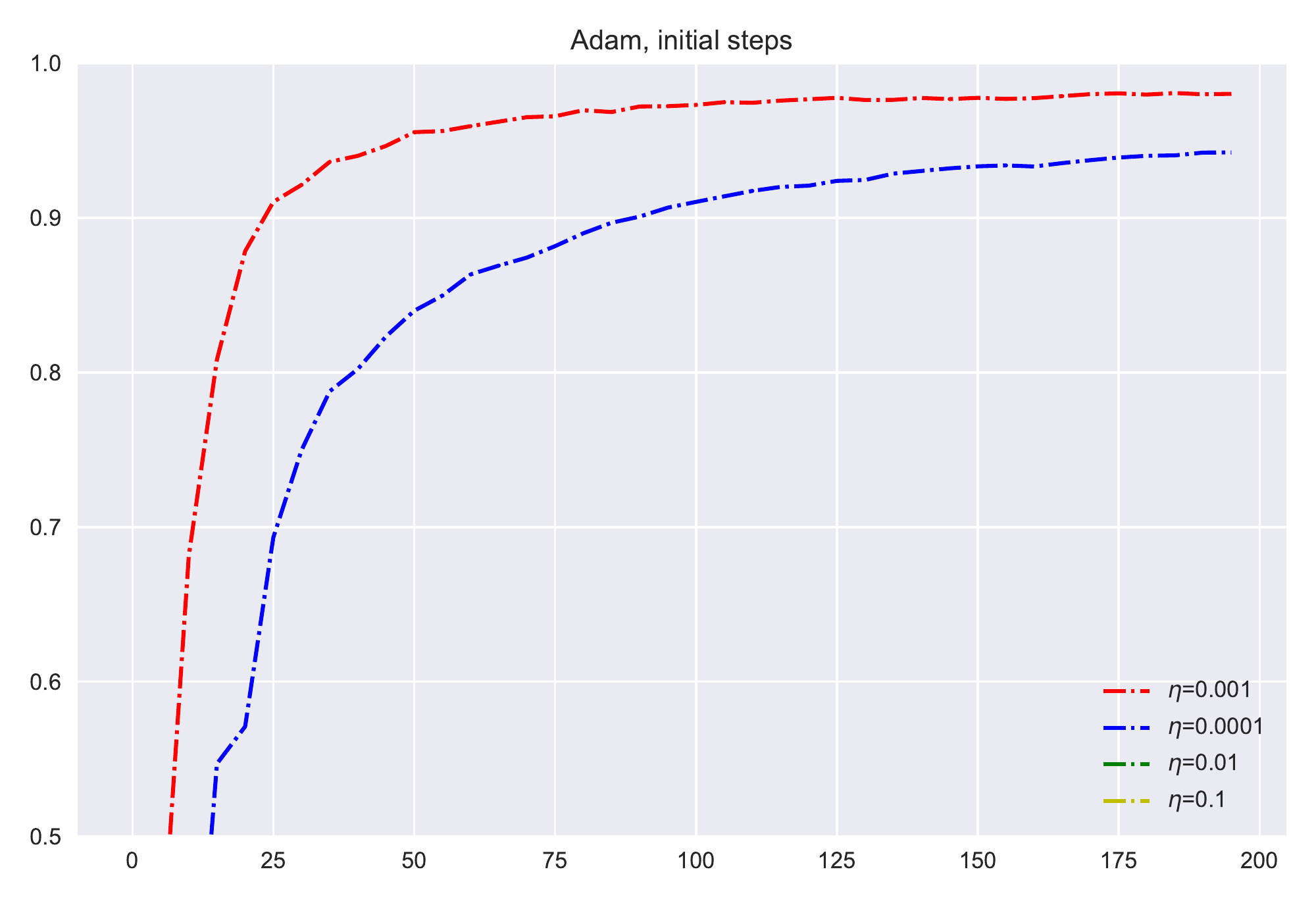}  & 	\includegraphics[width=0.33\columnwidth]{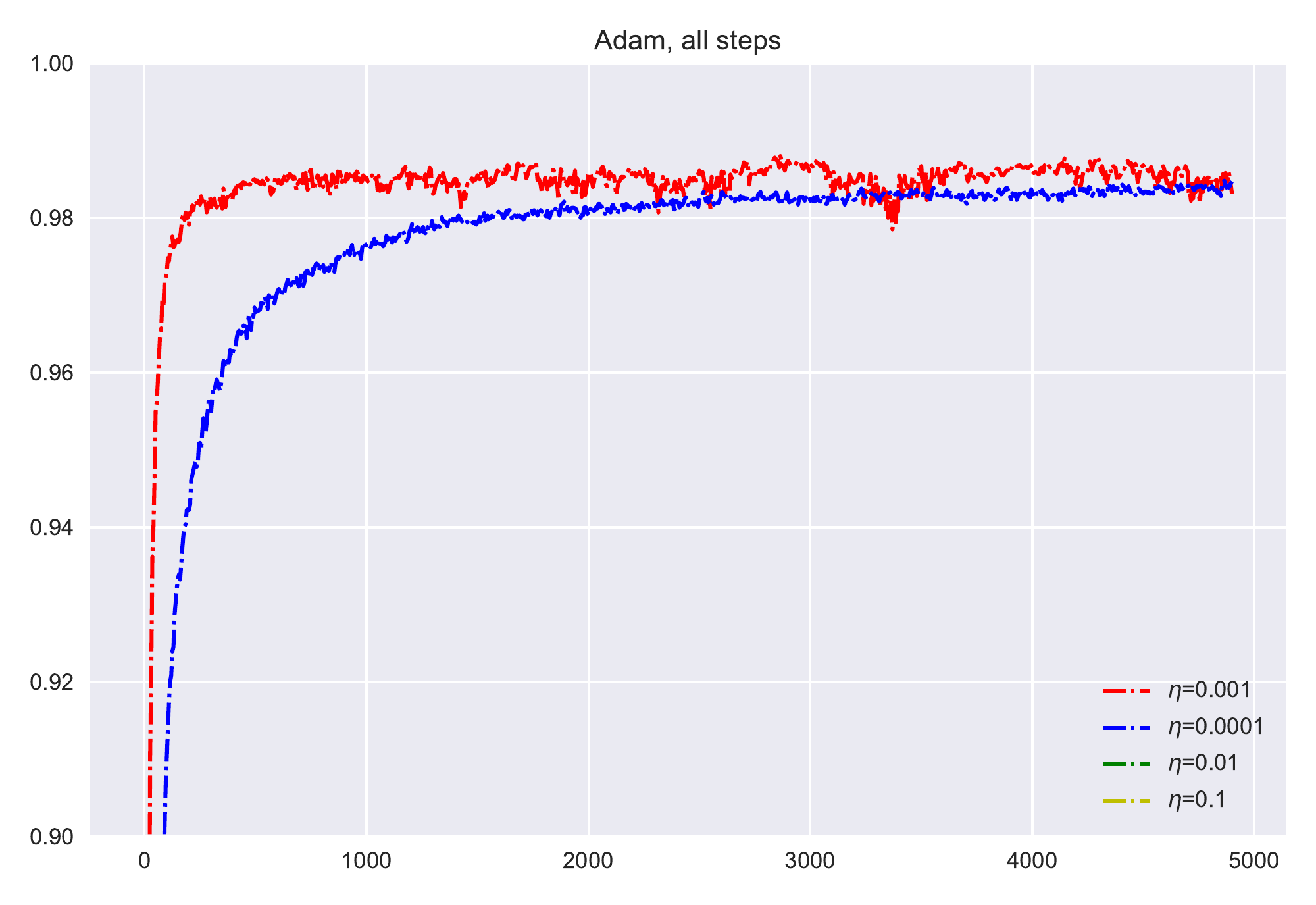}  &
	\includegraphics[width=0.33\columnwidth]{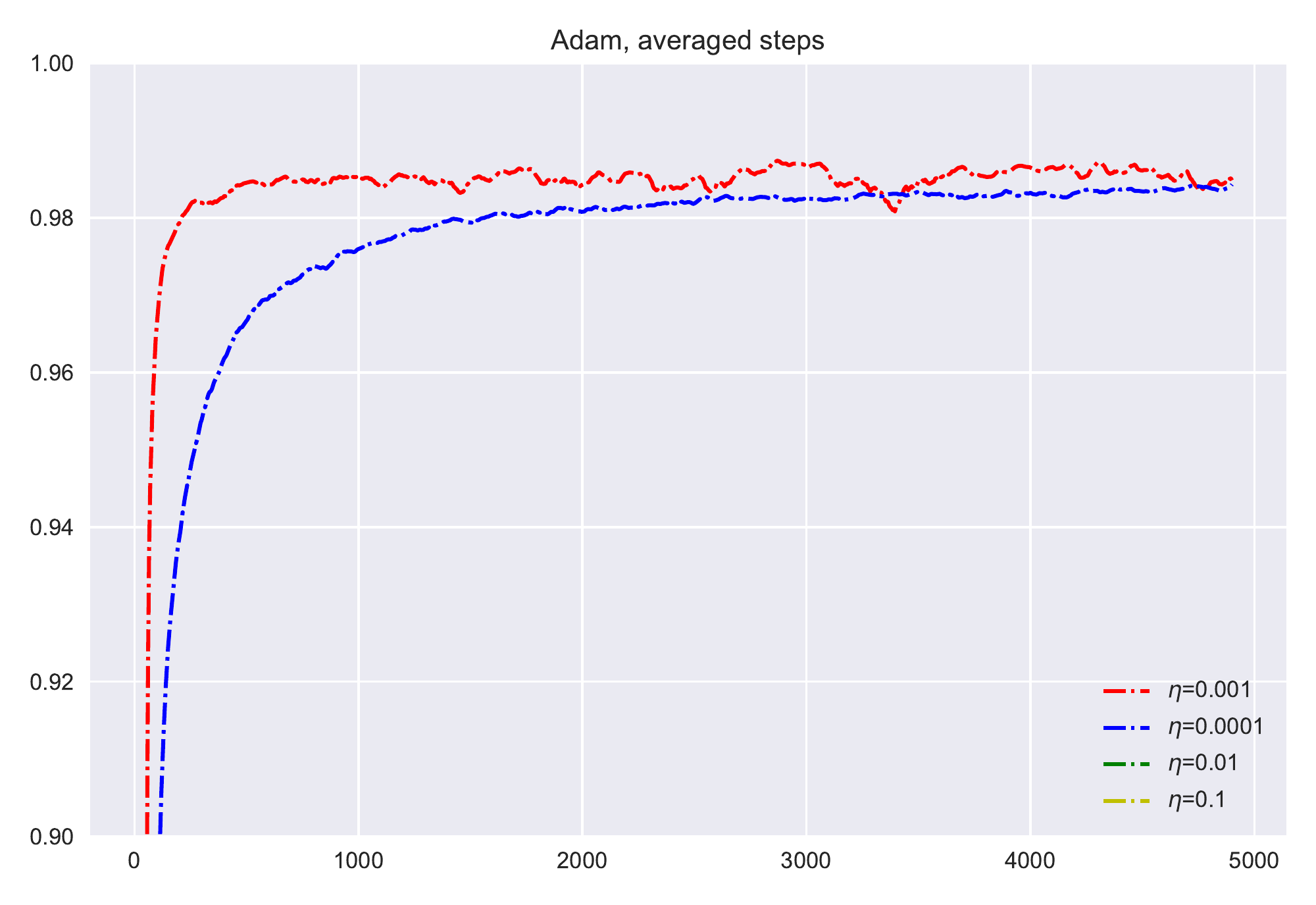}
	\\
	\includegraphics[width=0.33\columnwidth]{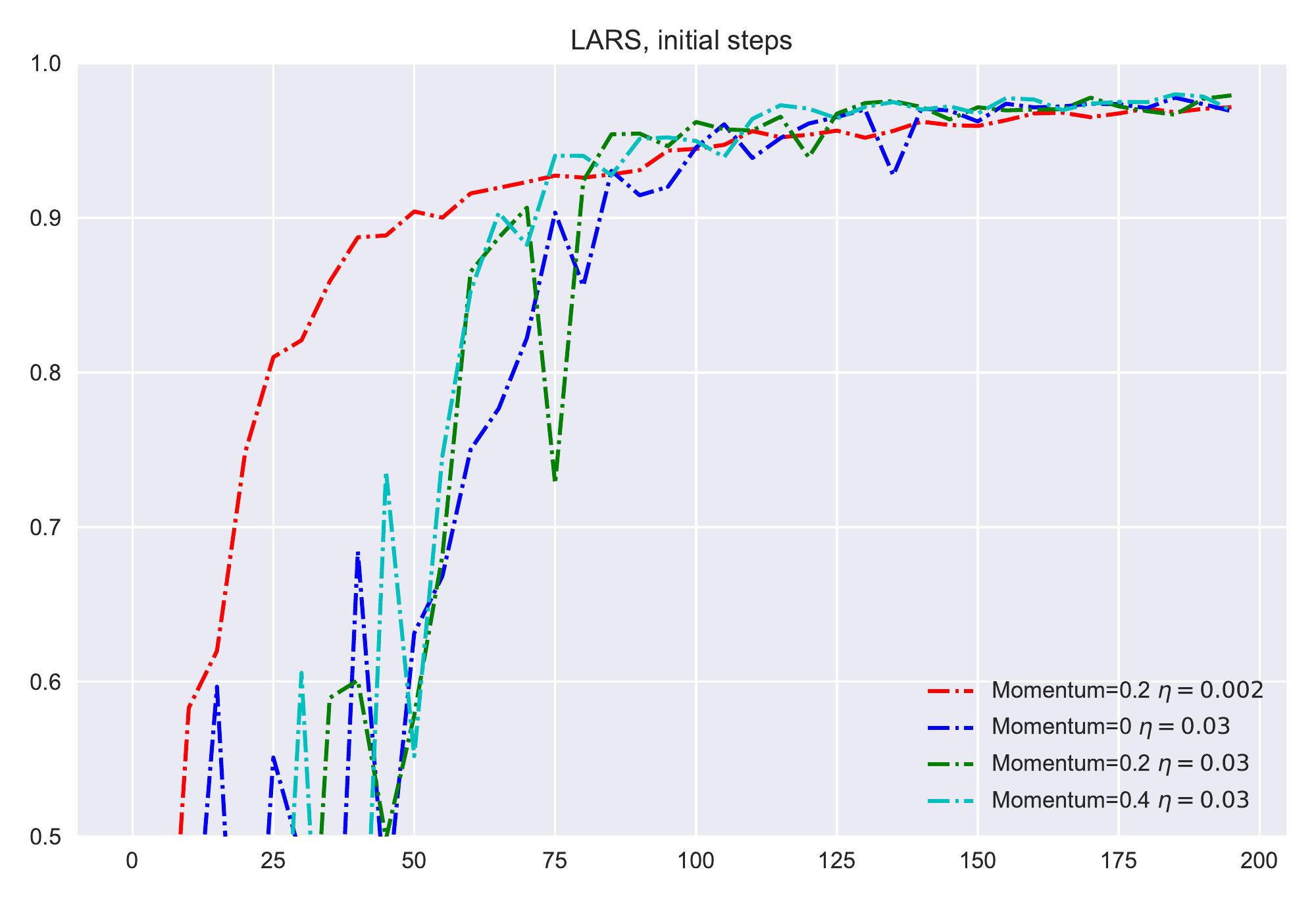}  & 	\includegraphics[width=0.33\columnwidth]{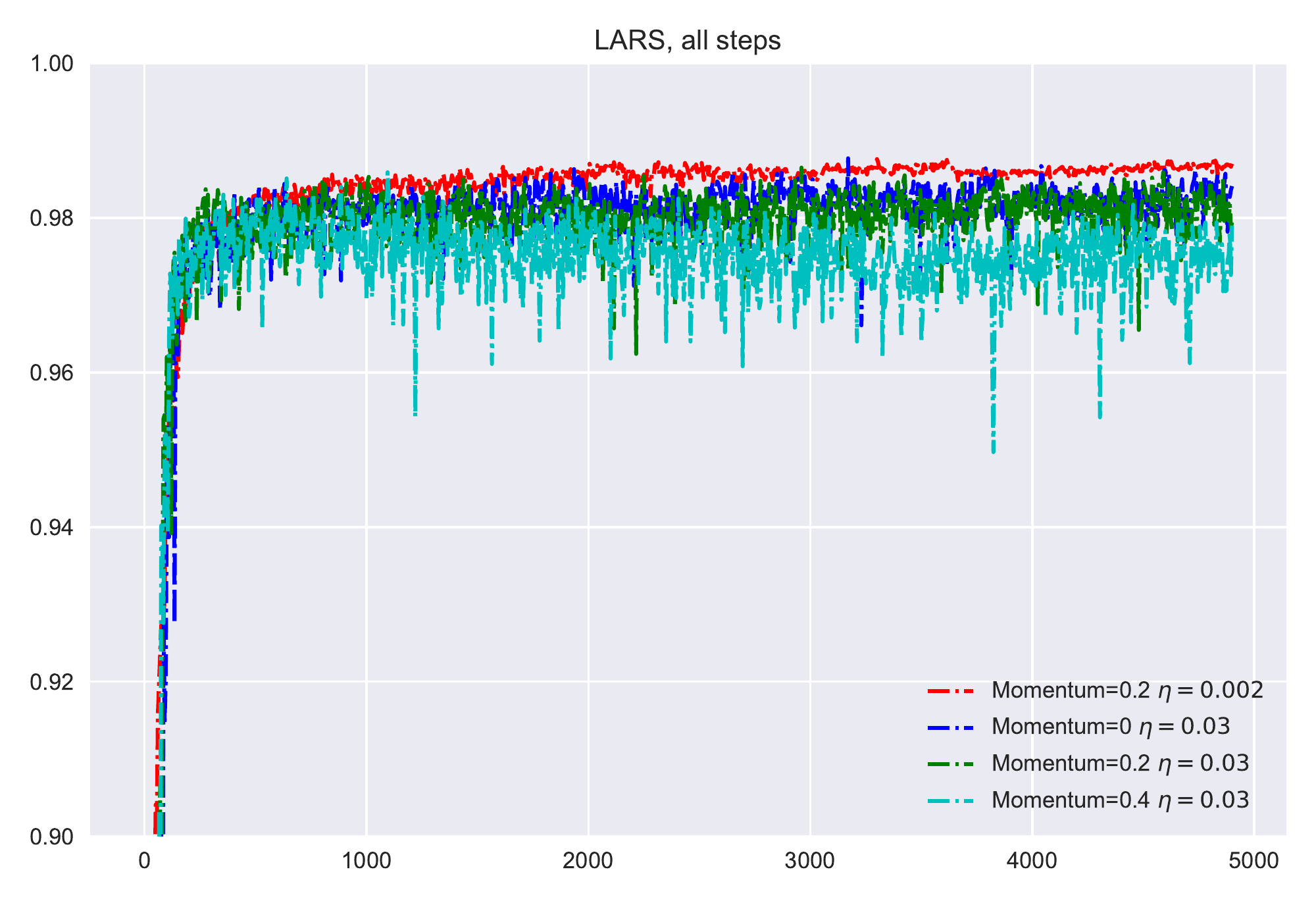}  &
	\includegraphics[width=0.33\columnwidth]{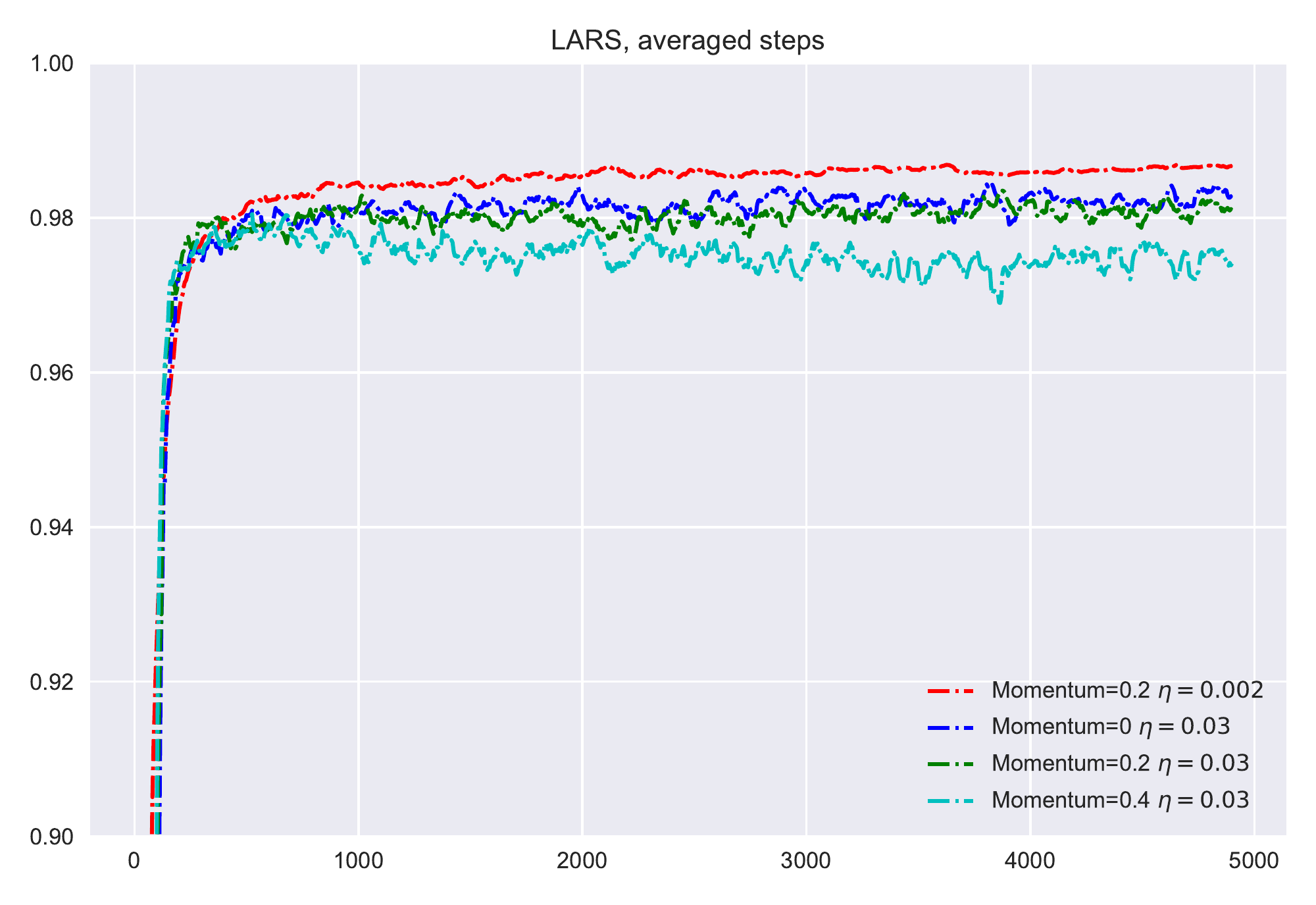}
	\\
	\includegraphics[width=0.33\columnwidth]{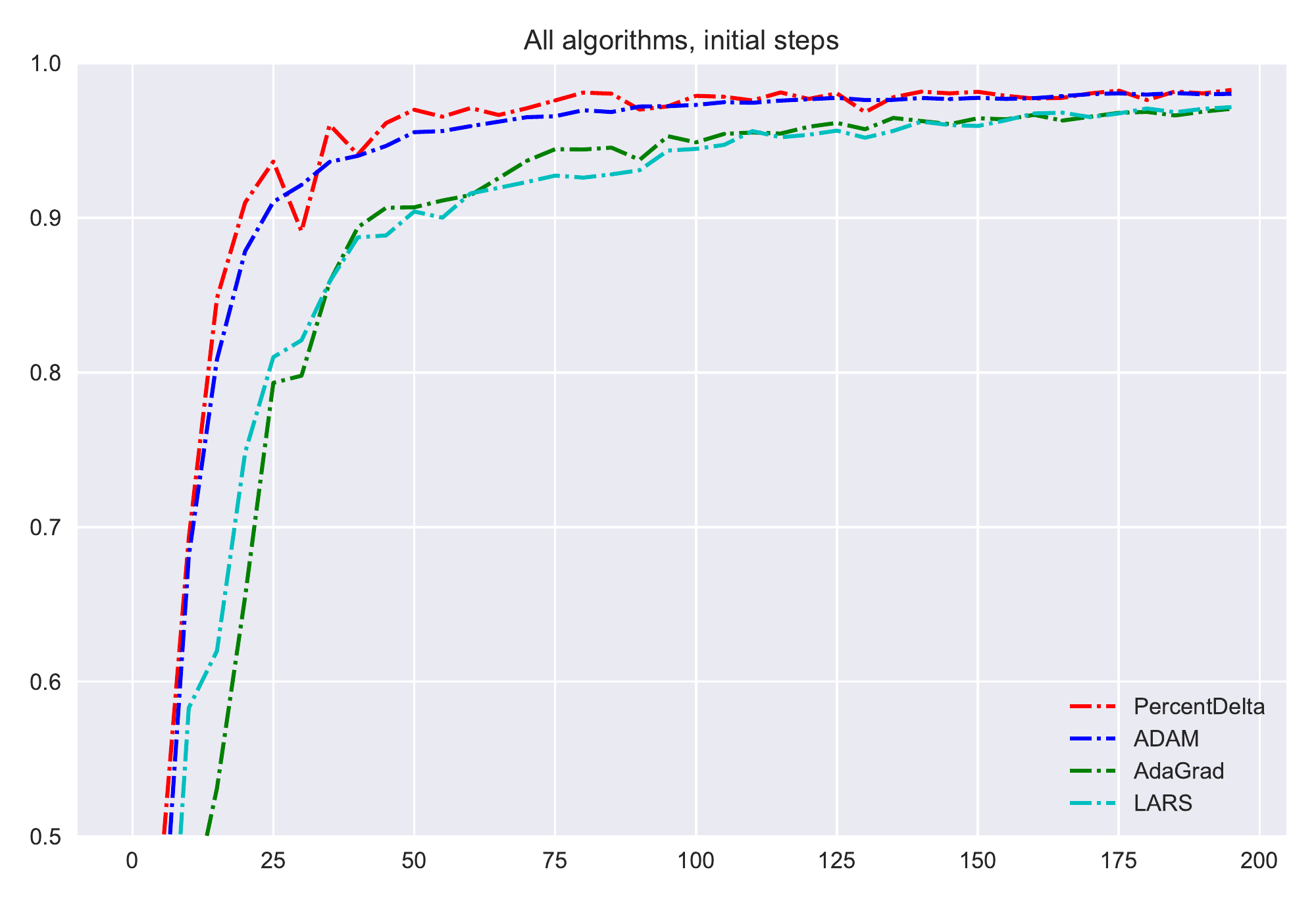}  & 	\includegraphics[width=0.33\columnwidth]{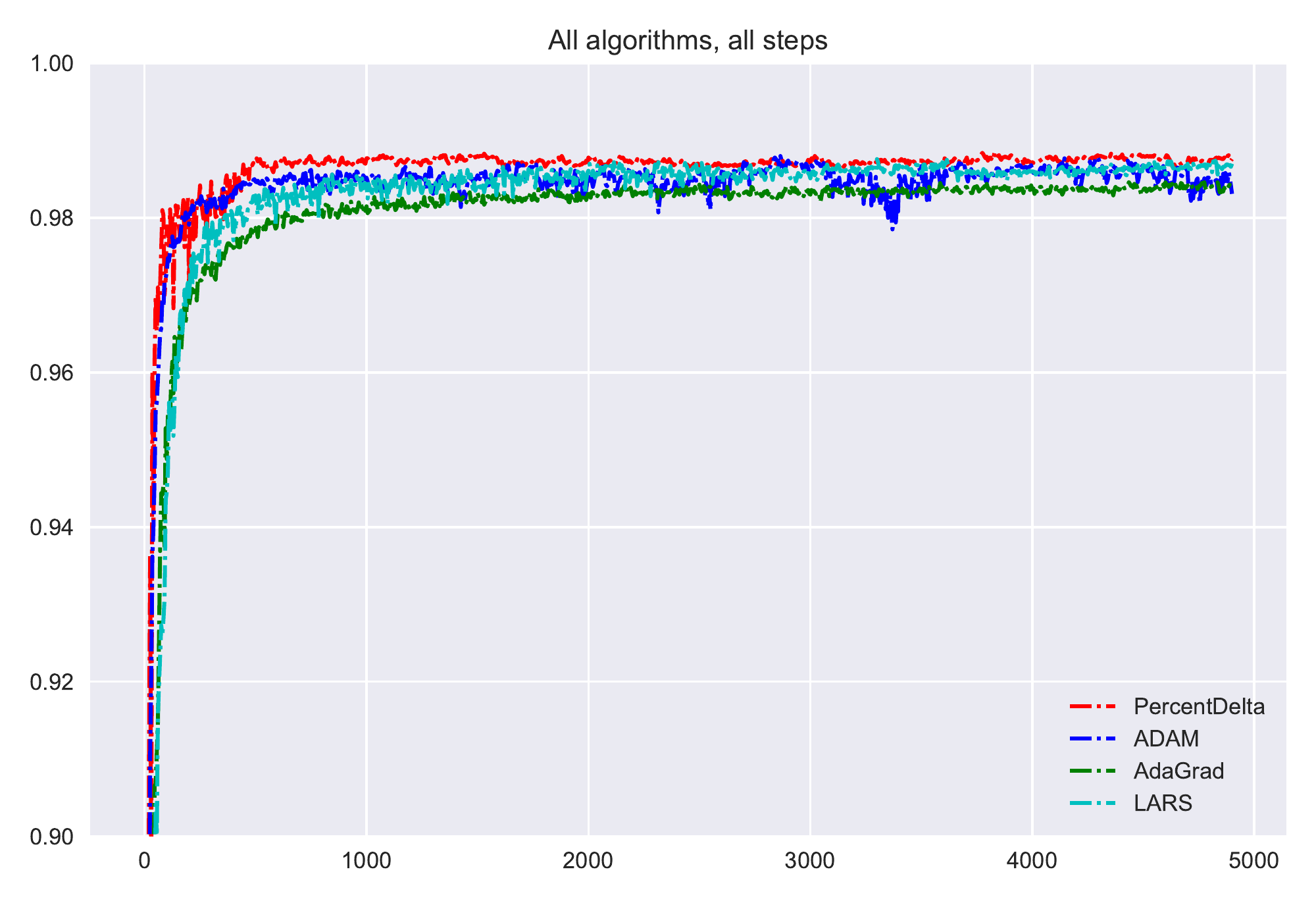}  &
	\includegraphics[width=0.33\columnwidth]{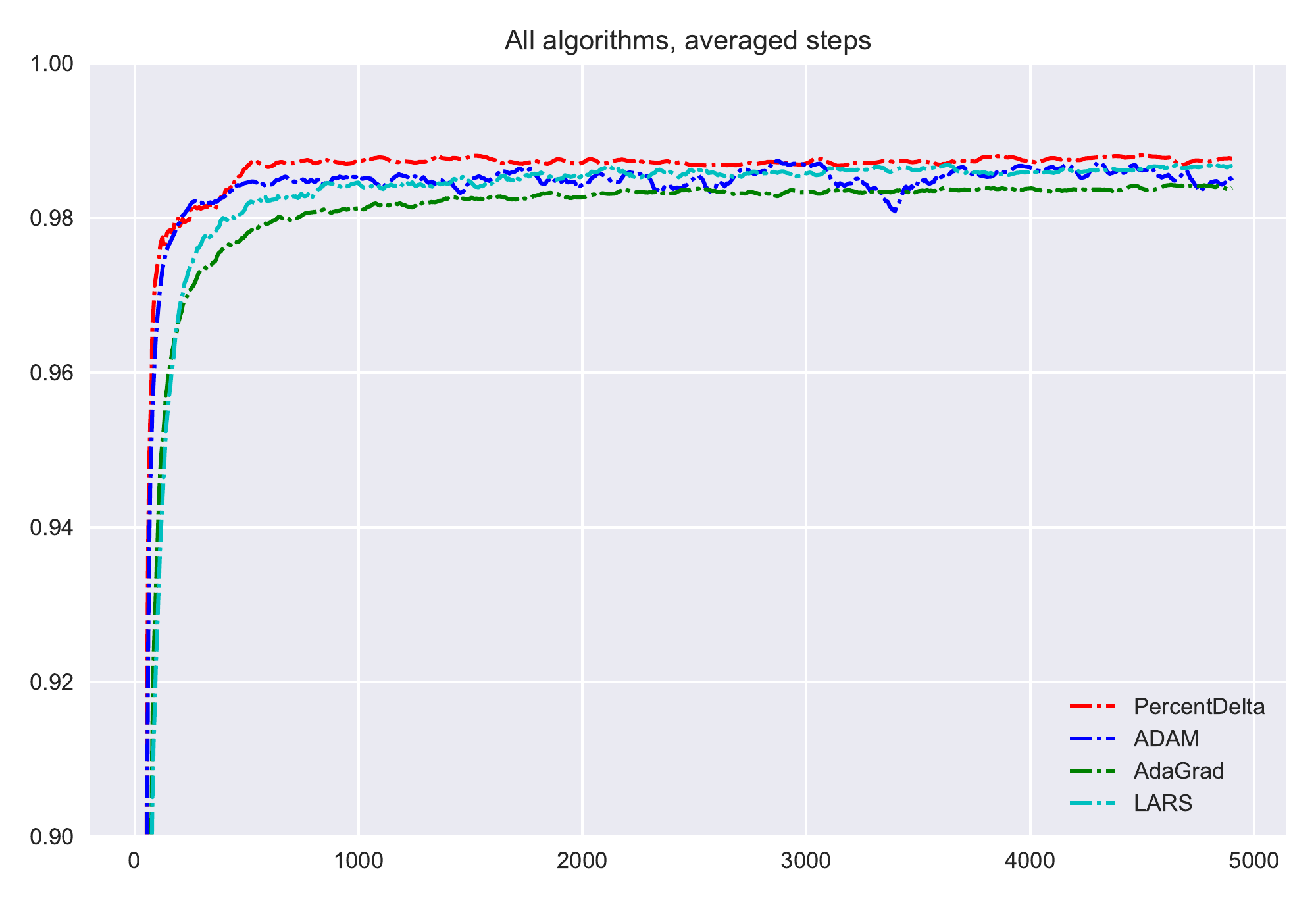}
	\\
\end{tabular}
\label{fig:mnist}
\caption{MNIST experiments -- Accuracy on Test set VS number of training steps. Each training step processes a batch size of 500 examples. The test accuracy is recorded after every 5-th steps. Figure contains subplots organized in 3 columns and 5 rows. \textbf{Columns} indicate the training stage. Left column shows early training (initial 200 steps) with y-axis starting at 0.5. Middle and right column show entire training curve, respectively showing raw values and smoothed values. \textbf{Rows} indicate algorithms. Each line color is consistent across the row and refers to the same hyper-parameter setting. We color the best performer with \textbf{red}. Top-to-bottom: PercentDelta (our algorithm), AdaGrad \citep{adagrad}, Adam \citep{adam}, LARS \citep{lars}, and the last row shows the best performer from all algorithms.}
\end{figure*}

Figure \ref{fig:mnist} 
In our experiments, we fix the learning rate of PercentDelta to $0.03$, use momentum, and set $\gamma(t)$ to Equation \ref{eq:gamma_m}. For Adagrad and Adam, we sweep the learning rate but we fix $\gamma(t) = 1$, as they implement their own decay. We do not use Momentum for Adagrad or Adam, as the later applies its own momentum scheme. For LARS, we use momentum, vary the learning rate, and set $\gamma(t)$ to Equation \ref{eq:gamma_m} with $m=0.01$. We note the following:
\begin{enumerate}
	\item In early stages, MNIST test accuracy climbs up the fastest with our algorithm.
	\item The final test accuracy produced by our algorithm is higher, given the training budget of 5000 steps.
\end{enumerate}

\section{Discussion}
\label{sec:discussion}
\subsection{Situations where PercentDelta is useful}
While PercentDelta has outperforms other training algorithms on 4-layer MNIST, the space of models-datasets is enormous we leave it as future work to try PercentDelta under various models and datasets. Nonetheless, we speculate that PercentDelta (and similarily, LARS, \cite{lars}) would be very useful in the following scenarios:
\begin{enumerate}
	\item Learning Embeddings. Consider the common setup of feeding word embeddings \citep[or graph embeddings,][]{asymproj} into a shared Neural Network and jointly learning the embeddings and Neural Network for an upstream objective. In this setup, a certain embedding vector is only affected by a fraction of training examples, while the shared network parameters are affected by all training examples. The sum of gradients w.r.t. the shared network parameters over all training examples can be disproportionately larger than embedding gradients. PercentDelta ensures that the shared network is not being updated much faster than the emebddings.

	\item Soft-Attention Models on Bag-of-Words. It is common to convert from variable-length bag-of-words $(\mathbf{x}_j[1], \mathbf{x}_j[2], \dots)$ into fixed-length representation by a convex combination: $\mathbf{x}_{j+1} := \sum_i \alpha_i \mathbf{x}_j[i]$, which can then be used for an upstream objective \cite[e.g. event detection in videos, ][]{ramanathan-cvpr16}. Here, $\alpha_i$ can be the $i$-th position of the softmax over all Words. The parameters of the softmax model would receive gradients from all words. PercentDelta ensures that, the otherwise disproportionately large, gradient updates of the softmax model are proportional to the remainder of the network.

	\item Matrix Factorization Models. For example, \cite{kore-bell} propose to factorize a user-movie rating matrix $R \in \mathbb{R}^{u \times m}$ into:\[
	R \approx W_U \times W_M + \mathbf{b}_U \times \vec{\mathbf{1}}^T + \vec{\mathbf{1}} \times \mathbf{b}_M + b,
	\]
	where $W_U \in \mathbb{R}^{u \times d}$ and $W_M \in \mathbb{R}^{d \times m}$ are the user and movie embedding matrix; $d$ is the size of the latent-space; $\mathbf{b}_U \in \mathbb{R}^u$ and $\mathbf{b}_M \in \mathbb{R}^m$ are the user and movie bias vectors, and $b \in \mathbb{R}$ is the global bias scalar. In this setup, $b$ would receive very large sum-of-gradients, and PercentDelta can ensure that all parameters are training at the same speed.
\end{enumerate} 

\subsection{Hyperparamters and Decay Function}
It seems that PercentDelta has many knobs to tune. However, we can fix $\eta$ to some value and only change $\gamma(t)$ as their product determines the effective rate of change across all layers' trainable tensors. We can set $\gamma(t)$ to constant decay:
\begin{equation}
\label{eq:gamma}
\gamma(t) = 1 - t \times m,
\end{equation}
where $0 < m << 1$ determines the decay slope. In addition, we can ensure that $\gamma(t) > 0$ to allow training continue indefinitely, by modifying Equation \ref{eq:gamma} to:
\begin{equation}
\label{eq:gamma_m}
\gamma(t) = \max(\beta, 1 - t \times m),
\end{equation}
where $\beta$ can be set to a small positive value, such as 0.01. In this case, if we fix $\eta = 0.03$, then we are effectively changing each trainable tensor by $3\%$ for every training batch initially, then gradually annealing this change-rate to $0.03\%$ after $\frac{1}{m}$ steps.

More importantly, we feel that $\eta$ and $\gamma(t)$ are a function of the dataset, and not the model. Experimentally, we observe the algorithm is insensitive to the choices of $m$ and $\eta$ as long as they are ``reasonable'' (i.e. removing diverging setups that can be quickly detected). However, we do not yet have a formula to automatically set them. Nonetheless, with a wide range of $\eta$ and $m$, we experimentally show on MNIST that PercentDelta beats all training algorithms, given the same budget of training steps.

\section{Conclusion}
\label{sec:conclusion}
We propose an algorithm that trains layers of a neural network, all at the same speed. Our algorithm, PercentDelta, is a simple modification over standard Gradient Descent. It divides the gradient w.r.t. a trainable tensor over the mean of $||\textrm{gradient / tensor}||_1$. The division over mean L1-norm is scalar, and only changes the gradient's magnitude but not its direction. Effectively, this updates the L1 norm of trainable layers, all at the same rate. We recommend a linear decaying change-rate schedule. Our modified gradients can be passed through a standard momentum accumulator \citep{momentum}. Overall, we show experimentally that our algorithm puts an upper envelop on all training algorithms, reaching higher test accuracy with fewer steps.
%In addition, we experimentally and analytically show that the gradient updates are relative to the tensor values.
%\newpage
%\input{appendix.tex}

\bibliographystyle{apalike}  % 
\bibliography{pd}

\end{document}